\documentclass[conference]{IEEEtran}
\usepackage[utf8]{inputenc}
\usepackage{graphicx}
\usepackage{amsmath,amssymb}
\usepackage{cite}
\usepackage{url}
\usepackage{multirow}
\usepackage{caption}
\usepackage{subcaption}
\usepackage{hyperref}
\usepackage{algorithm}
\usepackage{algpseudocode}
\usepackage{array} 
\usepackage{booktabs}
\usepackage[utf8]{inputenc}
\usepackage{placeins}
\usepackage{cleveref}
\usepackage{breakurl}  
\title{
Robotic System with AI for Real Time Weed Detection, Canopy Aware Spraying, and Droplet Pattern Evaluation
}

\author{
    Inayat Rasool$^{1}$,
    Pappu Kumar Yadav$^{1*}$, 
    Amee Parmar$^{1}$, 
    Hasan Mirzakhaninafchi$^{1}$, \\
    Rikesh Budhathoki$^{1}$,
    Zain Ul Abideen Usmani$^{1}$, 
    Supriya Paudel$^{1}$,  
    Ivan Perez Olivera$^{1}$,
    Eric Jones$^{2}$
    
    \\
    $^1$Machine Vision and Optical Sensor (MVOS) Lab, Dept. of Agricultural and Biosystems Engineering,\\
    South Dakota State University, Brookings, SD, USA \\
    $^2$Department of Agronomy, Horticulture and Plant Science, South Dakota State University, Brookings, SD, USA
    \\
    *Corresponding author: \texttt{pappu.yadav@sdstate.edu}
}

\begin{document}
\maketitle
\begin{abstract}
Uniform and excessive herbicide application in modern agriculture leads to increased input costs, environmental contamination, and rising herbicide resistance. To address these challenges, we developed a vision guided, AI-driven variable rate sprayer system capable of detecting weed presence, estimating canopy size, and adjusting nozzle activation in real time. The system uses lightweight YOLO11n and YOLO11n-seg models deployed on an NVIDIA Jetson Orin Nano for onboard inference and uses an Arduino Uno-based relay interface to control solenoid actuated nozzles based on canopy segmentation. Indoor trials were conducted using 15 potted \textit{Hibiscus rosa sinensis} plants of varying canopy sizes to simulate heterogeneous weed patches. The YOLO11n model achieved a mAP@50 of 0.98, precision of 0.99, and recall near 1.0. The segmentation model (YOLO11n-seg) achieved a mAP@50 of 0.48, precision of 0.55, and recall of 0.52. System performance was validated using water sensitive papers, which showed an average spray coverage of 24.22\% in canopy present zones. A clear upward trend in mean spray coverage from 16.22\% in small canopies to 21.46\% and 21.65\% in medium and large canopies demonstrated the system’s ability to dynamically modulate spray intensity in response to canopy size. These results highlight the feasibility of integrating real time deep learning with low-cost embedded hardware for selective herbicide application. Future work will expand detection to include three common South Dakota weed species i.e., waterhemp (\textit{Amaranthus tuberculatus}), kochia (\textit{Bassia scoparia}), and foxtail (\textit{Setaria spp.}) and validate the system further through both indoor and field trials in soybean and corn cropping systems.
\end{abstract}

\begin{IEEEkeywords}
real time weed detection, canopy aware spraying, variable rate spraying, spray pattern evaluation, robotic weed control system
\end{IEEEkeywords}

\section{Introduction}

Weed management remains one of the most persistent challenges in modern crop production. Weeds compete aggressively with crops for sunlight, nutrients, and water, and if left unmanaged, can cause significant yield losses often exceeding 30\% in major crops such as corn, soybean, and wheat \cite{lopez2011weed}. Conventional management methods predominantly rely on broadcast herbicide applications, which often result in inefficient chemical use, higher costs, and substantial environmental risks, including contamination of non-target areas and herbicide resistance buildup.

To reduce these negative impacts, site-specific weed management (SSWM) strategies have gained attention. SSWM focuses on detecting and treating only weed infested zones, significantly improving herbicide efficiency \cite{berge2012towards}. The integration of robotics and variable rate technology (VRT) offers a promising pathway to implement SSWM in real time agricultural settings \cite{upadhyay2024advances}. Ground-based robotic systems are particularly suited for such applications as they can carry larger payloads, maneuver close to canopy layers, and integrate with intelligent spraying mechanisms for precision targeting \cite{arakeri2017computer}.

Recent advancements in deep learning (DL) and computer vision (CV) have led to major breakthroughs in weed detection \cite{wang2007realtime,junior2021realtime,badhan2021realtime}. Real time object detection models such as YOLO have demonstrated high accuracy and fast inference speeds in complex field environments \cite{ferreira2017weed, razfar2022weed}. For instance, Yadav et al.\ (2023) applied YOLOv8 on UAV-based imagery to detect volunteer cotton plants in corn fields, demonstrating the model’s suitability for identifying weed like structures in crop rows \cite{yadav2023volunteer}. Similarly, Yadav et al.\ (2022) evaluated YOLOv5 across three different growth stages, highlighting its consistent performance under dynamic field conditions \cite{yadav2022yolov5}. Studies comparing YOLO variants and transformer-based architectures like RT-DETR have further expanded possibilities for robust, real-time weed identification under variable lighting and background conditions \cite{saltik2025comparative}. However, translating accurate weed detection into actionable spraying control remains a significant challenge.

Several studies have highlighted this gap, where detection is often decoupled from spraying mechanisms, or where spraying is binary (on/off) without consideration of canopy structure, density, or plant size \cite{dammer2007sensor, dammer2016realtime}. While high value crops like orchards have seen some progress with volume-based or height-based variable spraying, such approaches remain underexplored in row crops like soybean and corn \cite{luo2024extraction, salas2024design}.

Moreover, few studies comprehensively assess the physical efficacy of the spray itself, factors such as droplet size distribution, spray coverage, and spatial uniformity are often overlooked. Yet these are critical to determining field level effectiveness and minimizing off target deposition. Factors such as nozzle type, sprayer height, operating pressure, and environmental conditions greatly influence these spray characteristics \cite{etheridge1999characterization, martin2019effect, martin2024spray}.

Efforts to address these limitations are beginning to emerge. For instance, Partel et al. \cite{partel2019development} demonstrated a low cost AI-driven robotic sprayer that achieved high weed detection accuracy and precise localized spraying. However, such systems often require expensive Graphics Processing Units (GPUs) or complex integration frameworks, limiting their accessibility to small and medium scale growers. Similarly, Upadhyay et al. \cite{upadhyay2024advances} outlined the technological components and research gaps in SSWM, emphasizing the need for modular, scalable robotic platforms that integrate weed identification with targeted actuation.

In this study, we present the development and evaluation of a novel robotic spraying system that integrates a lightweight deep learning model (YOLO11n) with real time canopy aware actuation. The system is mounted on a four nozzle boom and utilizes a custom-built controller with PWM regulated valves to adjust spray output based on segmented canopy area. We further evaluate the system's physical spraying performance using water sensitive paper (WSP) to quantify spray coverage and droplet size distribution.

The specific objectives of this study were to: (i) develop and deploy a YOLO11n-based real time weed detection model on a ground robotic platform; (ii) implement a canopy segmentation algorithm to guide variable spray actuation; and (iii) analyze spray coverage, droplet size, and spatial density using image analysis of water sensitive paper (WSP). By addressing both detection and actuation in a single system, this work aims to close the loop between AI-driven perception and physical weed control for practical, scalable SSWM solutions.

\section{Material and Methods}
\subsection{Sprayer Assembly on Ground-Based Robotic System}

The variable rate sprayer was integrated onto a Farm-ng Amiga (Farm-ng, Inc., Watsonville, CA, USA) platform, a fully electric, modular skid-steer robot designed for field research and precision agriculture. The Amiga chassis is constructed with powder-coated mild steel forks and frame components, weighing approximately 150 kg and supporting a rated payload of 450 kg, sufficient to accommodate tanks, control units, and embedded processors \cite{farmng2024docs}.

The platform is powered by four sealed brushless DC geared motors, each rated between 250 to 500 W with a 1:30 reduction ratio, and delivering up to 140 Nm of peak torque. Motors are controlled by CAN integrated field-oriented control (FOC) motor controllers and operate at 36 to 43 VDC through a 44 V nominal power circuit. To ensure thermal stability, a built-in safety mechanism halts motor operation once temperatures exceed 80°C. During system setup, the robot has wheelbase of 33 inches (84 cm), trackwidth of 60 inches (152 cm), and ground clearance of around 62 inches (157 cm) \cite{farmng2024docs}.

A custom aluminum spray boom was mounted at the rear of the robot using U clamps. The boom included four spray nozzle bodies spaced 30 cm apart, each fitted with a 12VDC PWM actuated solenoid valve. The liquid supply was delivered from a 50 liter tank mounted on the robot, pressurized using a diaphragm pump. Flow was regulated through an inline flow control valve. The nozzle actuation was controlled by an Arduino Mega (Arduino S.r.l., Somerville, MA, USA) receiving actuation commands serially from an NVIDIA Jetson Orin Nano, which processed real time image data from a front-facing RGB camera mounted on the spray boom.

The Amiga was equipped with the Farm-ng Intelligence Kit, including the Brain, a PoE switch, GPS antenna, and additional stereo cameras. The Brain was powered via the upper PoE switch port and connected to the CAN bus, while cameras received power from the lower ports. The GPS antenna provided RTK corrections and was connected using an SMA to TNC cable. Whenever the mount location of the IMU or GPS was adjusted, the geometry parameters of the robot were updated accordingly in the Brain settings.

\begin{figure*}[t]
    \centering
    \includegraphics[width=\textwidth]{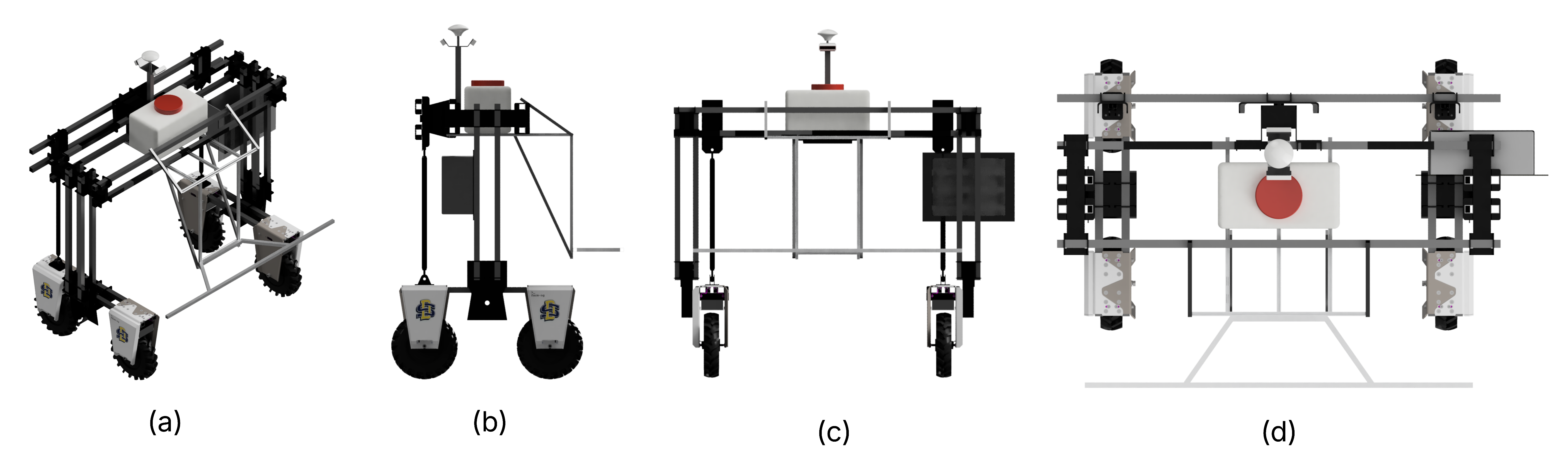}
    \caption{CAD model of the mechanical structure of the spraying system from multiple perspectives:(a)isometric, (b)side, (c)front, and (d)top views.}
    \label{fig:cadmodel}
\end{figure*}

\begin{figure*}[t]
    \centering
    \includegraphics[width=\textwidth]{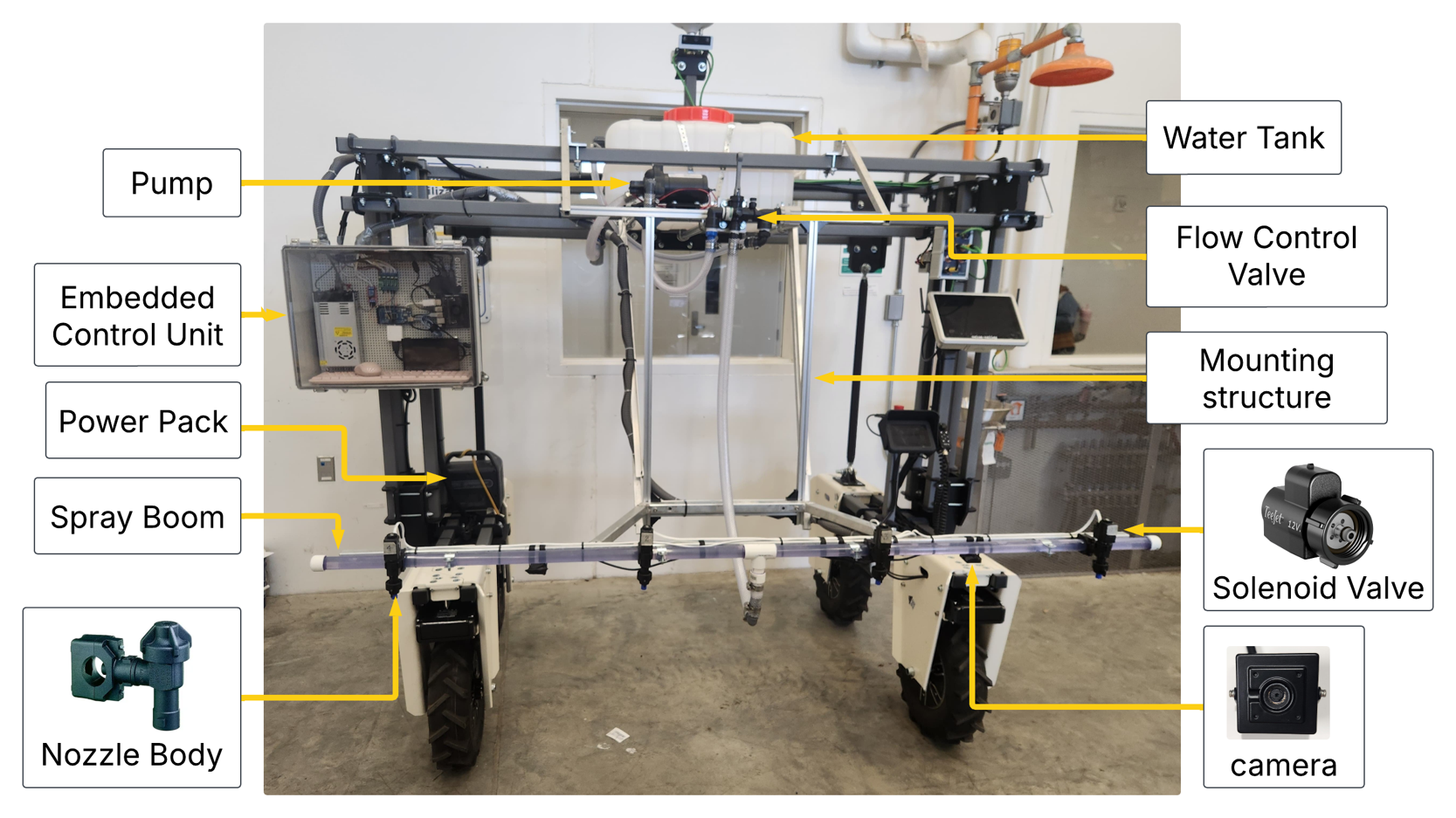}
    \caption{Custom variable-rate sprayer system mounted on the Farm-ng Amiga platform, showing the integrated water tank, flow control valve, pump, spray boom, solenoid valves, and camera used for real-time weed detection and targeted application.}
    \label{fig:sprayer_system}
\end{figure*}

The completed assembly of the robotic system is visualized in Figure ~\ref{fig:cadmodel}, showing the CAD model from multiple perspectives (isometric, side, front, and top views) to demonstrate component integration. The actual overall system is shown in Figure~\ref{fig:sprayer_system}, which illustrates the mechanical integration of all components including the embedded control unit, tank, pump, spray boom, solenoid valves, and front-mounted camera. The compact, robust layout enabled effective real-time weed detection and precision spraying during field operation.

\subsection{AI-driven Embedded Control Unit}

A dedicated embedded control unit was designed to manage the perception, decision making, and actuation tasks of the sprayer system in real time. Housed in a weather resistant enclosure mounted to the robot frame, the unit integrated all computing, communication, and power control components necessary to operate the system independently.

At the core of the unit was an NVIDIA Jetson Orin Nano board, which served as the primary computing device. It ran the deep learning-based weed detection model and handled image acquisition, inference, and communication with the microcontroller. A 7 inch touchscreen display was mounted inside the enclosure to provide a local interface for system monitoring and debugging. The display was connected via HDMI and powered through a regulated micro -USB port.

An Arduino Mega microcontroller acted as the intermediary between the Jetson and the spray system hardware. It received digital signals over USB serial from the Jetson and translated them into activation commands for the relay module. The relay module controlled four individual 12V PWM solenoid valves on the spray boom. To support fine-grained spray modulation, a PWM switching module was integrated downstream of the relay module, allowing spray durations to be adjusted based on weed canopy area.

The control unit also contained a 12V/20A regulated power supply connected to a fused terminal block, ensuring safe and distributed power to the solenoids, microcontroller, relay board, and switching module. All internal wiring was terminated using labeled screw terminals for ease of serviceability and modularity. The enclosure was designed with proper ventilation to prevent overheating during prolonged operation in outdoor field conditions.

\begin{figure}[htbp]
    \centering
    \includegraphics[width=0.48\textwidth, height=5.5cm]{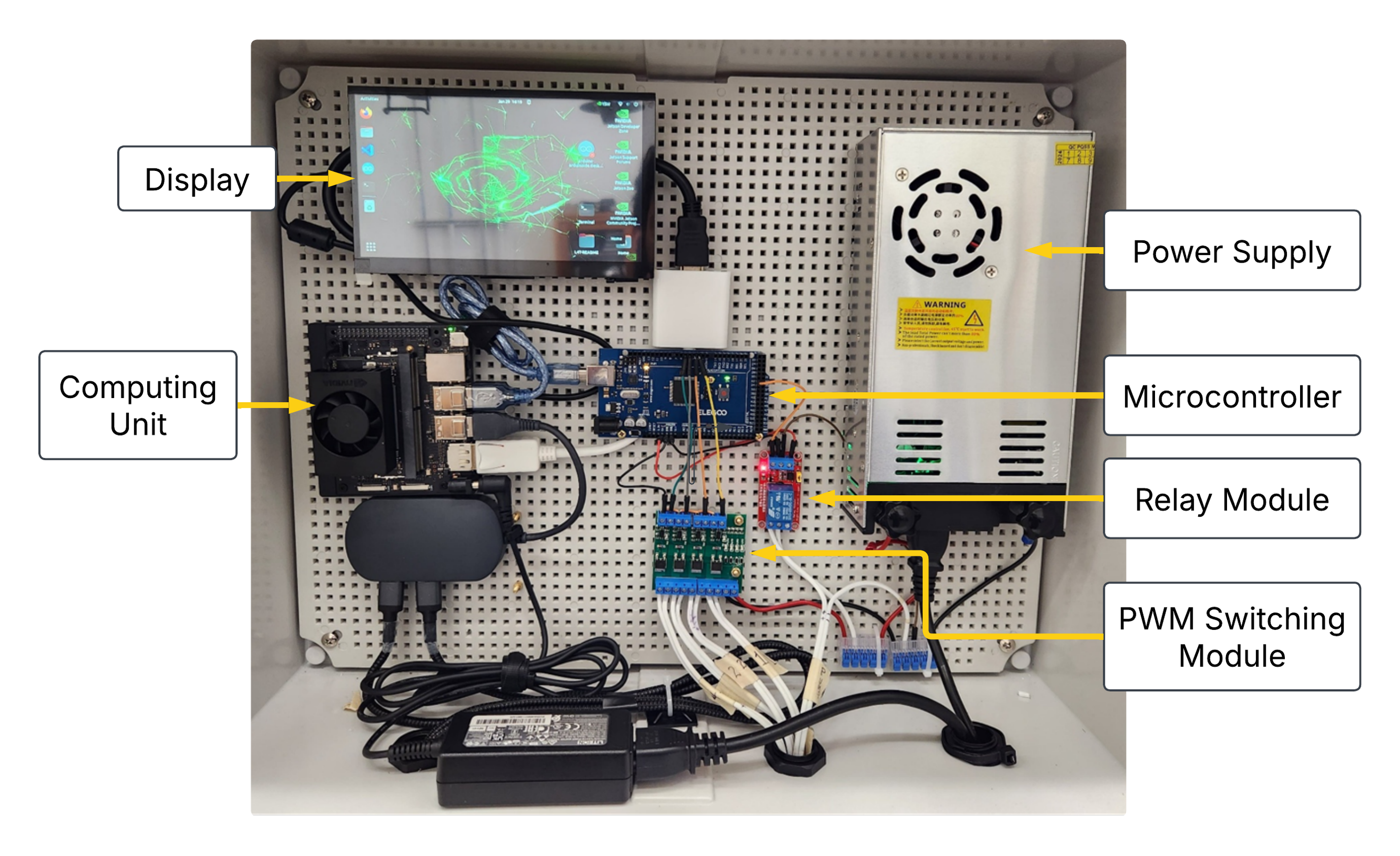}
    \caption{Embedded control unit showing the computing, microcontroller, power, and relay components used for real-time weed detection and spray actuation.}
    \label{fig:control_unit}
\end{figure}

A labeled schematic of the control unit showing the computing unit, microcontroller, power supply, relay modules, PWM module, and display is shown in Figure~\ref{fig:control_unit}.
  
\subsection{Spray Boom Assembly}

\begin{figure}[htbp]
    \centering
    \includegraphics[width=0.48\textwidth]{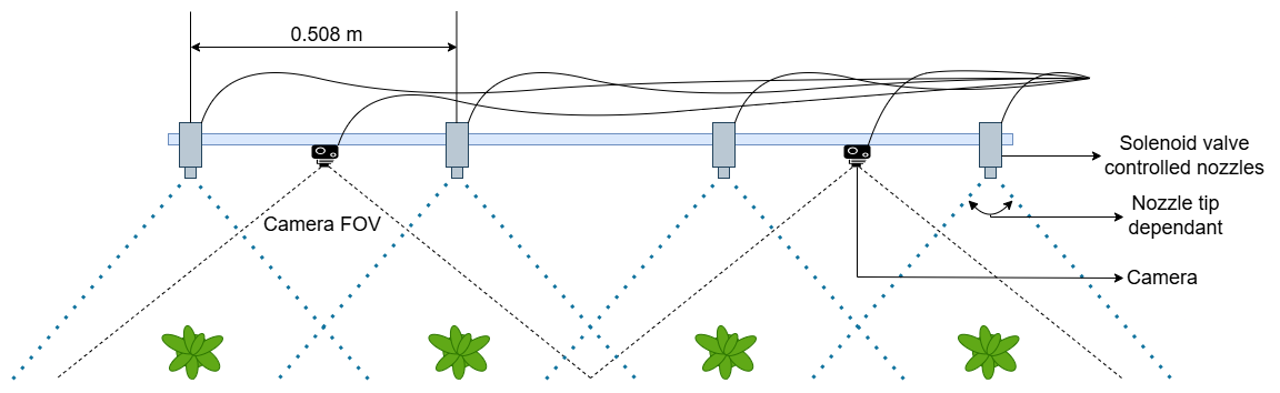}
    \caption{Conceptual layout of the spray boom showing 0.508 m nozzle spacing, camera positions, and nominal spray patterns.}
    \label{fig:boom_overview}
\end{figure}

The spray boom was designed to deliver herbicide precisely based on canopy coverage while minimizing overlap between adjacent spray patterns. A 1.5 meter aluminum wet boom (0.75 inch outer diameter) served both as a fluid conduit and a structural mounting rail. Four TeeJet QJ22187–¾ nozzle bodies (TeeJet Technologies, Springfield, IL, USA) were clamped at 0.5 meter intervals along the boom (Fig.~\ref{fig:boom_overview}). Each nozzle body was equipped with a TeeJet 115880-2-12 solenoid valve (12 V DC) (TeeJet Technologies, Springfield, IL, USA) to enable individual PWM control. These valves were actuated by signals from the embedded control unit (Figure~\ref{fig:control_unit}) and supported minimum pulse durations as short as 20 milliseconds, enabling real time adjustment of spray output.

At the outlet of each solenoid valve, a TeeJet TP80015EVS even flat fan tip (80° spray angle, 0.15 GPM at 40 psi) (TeeJet Technologies, Springfield, IL, USA) was installed to generate a uniform spray band, optimized for banded herbicide applications. The nozzle tips were selected to achieve complete ground coverage at a boom height of 0.35 meters positioned below canopy level to minimize drift. A low pressure 12 V diaphragm pump maintained a constant pressure of 40 psi at the boom inlet, with measured pressure losses of less than 2 psi under a total flow demand of 0.6 GPM.

Two Arducam USB cameras (Sony IMX219 sensor, 8 MP resolution, with 72° diagonal, 60° horizontal, and 47° vertical fields of view) (Arducam Technology Co., Ltd., Shenzhen, China) were mounted between alternating nozzle bodies. This configuration ensured that each camera covered the spray zones of two adjacent nozzles (Fig.~\ref{fig:boom_overview}), allowing targeted spray decisions within discrete sectors and avoiding overlap or double dosing.

All electrical connections were routed along the top of the boom within a flexible harness protected by harness wrap. Quick-disconnect Deutsch DT06 connectors enabled rapid field replacement of individual valves. Short service loops were included at each junction to prevent wire strain and accommodate boom flexing during operation on uneven terrain.

\begin{figure}[htbp]
\centering
\begin{subfigure}[b]{0.2\textwidth}
    \centering
    \includegraphics[width=\linewidth]{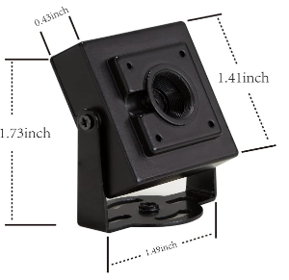}
    \caption{Arducam IMX219 camera \cite{arducamIMX219}} 
\end{subfigure}
\hfill
\begin{subfigure}[b]{0.2\textwidth}
    \centering
    \includegraphics[width=\linewidth]{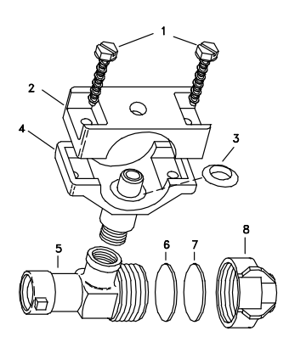}
    \caption{QJ22187 nozzle body \cite{teejetNozzleBody}}
\end{subfigure}
\hfill
\begin{subfigure}[b]{0.2\textwidth}
    \centering
    \includegraphics[width=\linewidth]{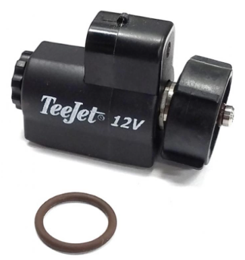}
    \caption{115880-2-12 solenoid \cite{teejetSolenoid}}
\end{subfigure}
\hfill
\begin{subfigure}[b]{0.2\textwidth}
    \centering
    \includegraphics[width=\linewidth]{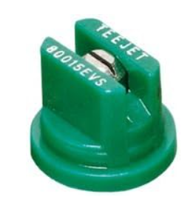}
    \caption{TP80015EVS tip \cite{teejetSprayTip}}
\end{subfigure}
\caption{Key components of the spray boom system.}
\label{fig:boom_parts}
\end{figure}

\subsection{Experiment Design for Indoor Trials}

\begin{figure}[htbp]
    \centering
    \includegraphics[width=0.48\textwidth]{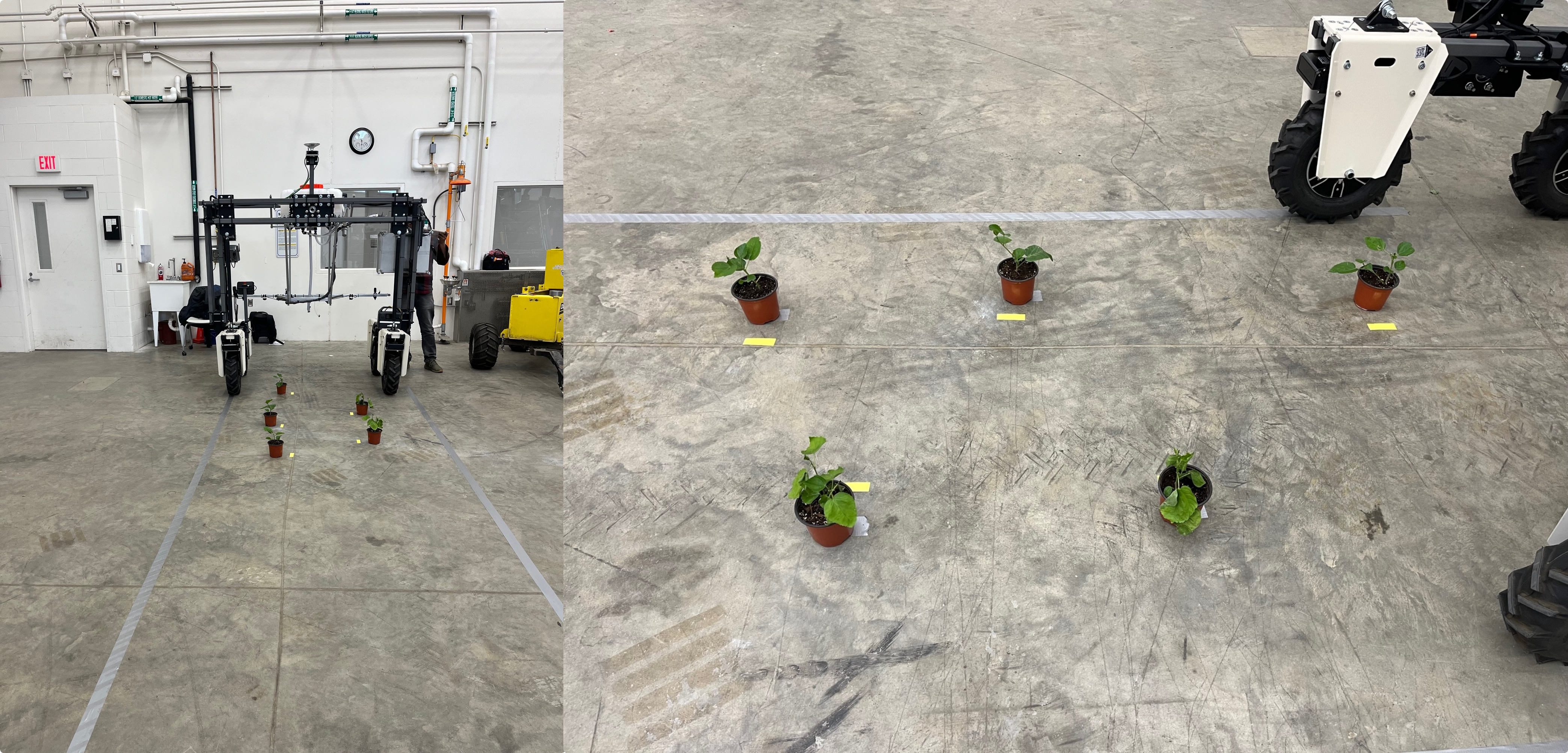}
    \caption{Indoor experiment trial showing potted hibiscus plants categorized by canopy size (small, medium, large).}
    \label{fig:exp_trial}
\end{figure}

To evaluate the performance of the vision guided precision spraying system in a controlled environment, an indoor experiment was conducted using fifteen potted hibiscus plants (\textit{Hibiscus rosa sinensis} L.) that were purchased from a local nursery in Brookings, South Dakota. Hibiscus plants were chosen due to their broadleaf structure, which closely resembles common weed morphology in row crop systems. The fifteen plants were intentionally selected to represent three different canopy size categories i.e., five with small canopy area, five with medium canopy area, and five with large canopy area (Fig.~\ref{fig:exp_trial}). This helped to introduce natural variation in plant structure for both vision based detection and variable rate spray actuation testing.

All 15 plants were used for both image acquisition and real time spraying evaluation. RGB images were captured under varying lighting conditions and from multiple angles to simulate the diversity encountered in real-world field scenarios. These images were used to train two deep learning models: YOLOv11n for object detection and YOLOv11n-seg for canopy segmentation. Both models were developed with deployment on the Jetson Orin Nano in mind, ensuring real-time performance on edge devices.

To improve model generalization, common data augmentation techniques such as rotation, flipping, scaling, and contrast/brightness adjustments were applied to the original image set. The augmented dataset was then divided into training, validation, and testing sets using an 80:10:10 ratio.

This controlled indoor setup provided a stable, repeatable testing environment free from external disturbances like wind or lighting shifts. It enabled fine-tuning of both the perception system and the sprayer actuation logic before field trials involving real weed populations in outdoor soybean plots.

\subsection{YOLO11 Network Architecture}

\begin{figure}[htbp]
    \centering
    \includegraphics[width=0.48\textwidth]{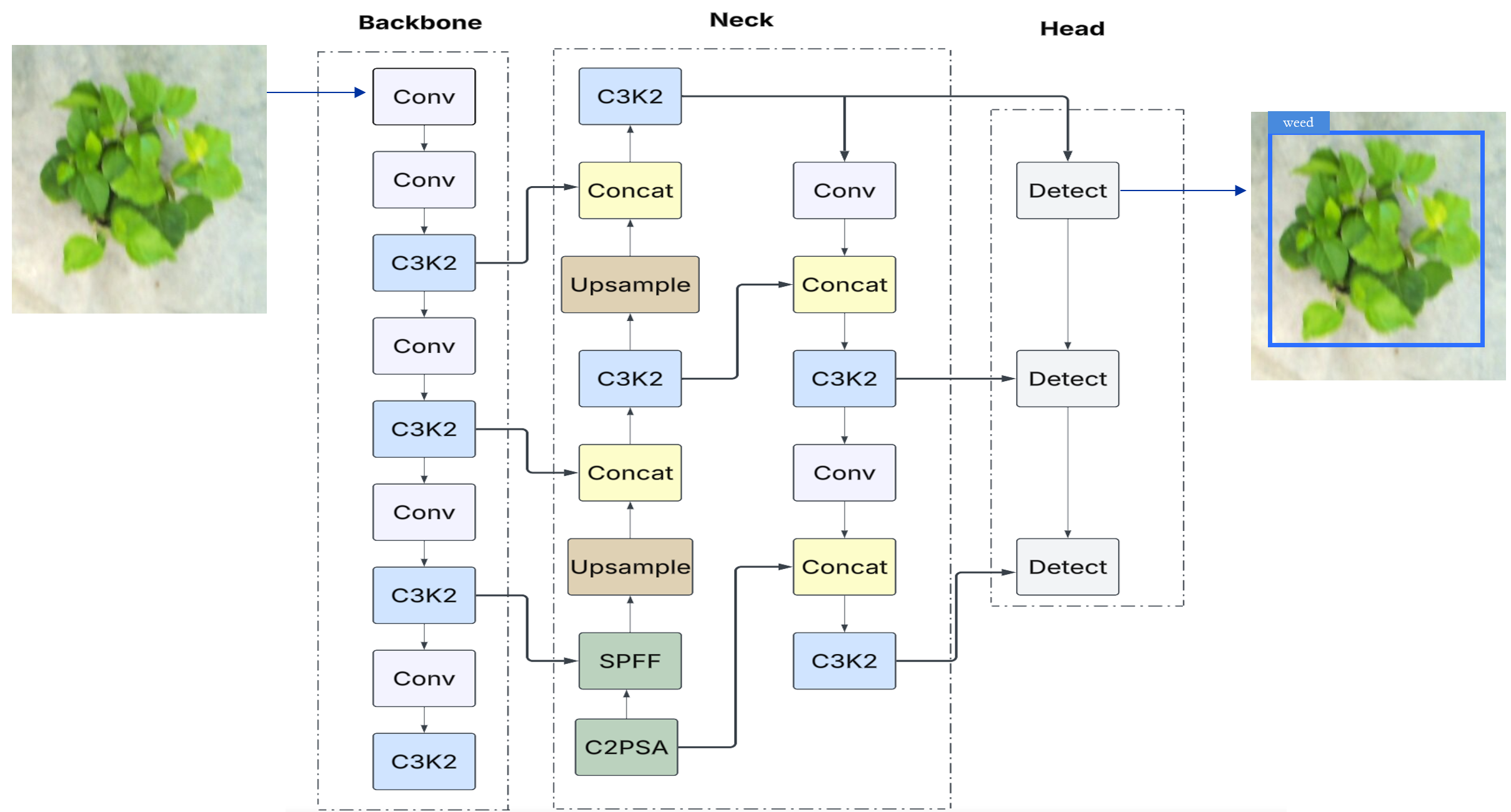}
    \caption{Schematic overview of the YOLO11 architecture used for real time weed detection (left) and the YOLO11n-seg extension for semantic segmentation (right).}
    \label{fig:yolo11_arch}
\end{figure}

The detection backbone in this study is the lightweight You Only Look Once version 11 (YOLO11) network, an evolution of the YOLOv5/YOLOv8 family that balances inference speed with high detection accuracy. YOLO11 adopts a CSP–Ghost backbone for efficient feature reuse, incorporates depthwise separable convolutions to reduce parameter count, and employs a decoupled head with separate branches for classification, bounding box regression, and objectness scores. For the segmentation variant (YOLO11n-seg), a lightweight decoder with panoptic feature aggregation is appended to generate pixel wise masks while sharing the majority of the detection backbone. The network was trained end-to-end with mosaic augmentation, CIoU loss for bounding boxes, and binary cross-entropy for mask predictions. Figure~\ref{fig:yolo11_arch} illustrates the high level architecture adopted in this work. Full implementation details and source code are available in the official repository \cite{yolo11github}.

\subsection{Performance Metrics and Training Configuration}

A total of 443 RGB images were collected and labeled for model development. The dataset was split into training, validation, and test sets in a 387:38:18 ratio, respectively. All training was carried out on South Dakota State University's High-Performance Computing (HPC) cluster called Innovator \cite{innovator2025}, which provides access to CPU and GPU nodes through the Simple Linux Utility for Resource Management (SLURM) workload manager. Training was performed on a GPU node equipped with an NVIDIA T4 GPU (NVIDIA Corporation, Santa Clara, CA, USA) and 48-core Intel Xeon CPUs.

The YOLO11n detector was trained for 400 epochs until convergence, while the YOLO11n-seg segmentation model converged in 267 epochs. The models were optimized using the Adam optimizer with a learning rate of 0.001, batch size of 16, and early stopping based on validation loss.

Model performance was assessed using class-level precision (\(P\)), recall (\(R\)), the harmonic \(F_{1}\) score, and mean average precision (mAP). Let \(\mathrm{TP}\), \(\mathrm{FP}\), and \(\mathrm{FN}\) denote true positives, false positives, and false negatives, respectively. Precision and recall are defined as:
\begin{equation}
P \;=\; \frac{\mathrm{TP}}{\mathrm{TP} + \mathrm{FP}}, 
\label{eq:precision}
\end{equation}
\begin{equation}
R \;=\; \frac{\mathrm{TP}}{\mathrm{TP} + \mathrm{FN}}.
\label{eq:recall}
\end{equation}
The \(F_{1}\) score combines \eqref{eq:precision} and \eqref{eq:recall}:
\begin{equation}
F_{1} \;=\; 2 \,\frac{P \times R}{P + R}.
\label{eq:f1}
\end{equation}
Detection accuracy across confidence thresholds is summarized by the average precision (\(\mathrm{AP}\)) for each class; the mean over all classes yields mAP:
\begin{equation}
\text{mAP} \;=\; \frac{1}{C}\sum_{c=1}^{C} \mathrm{AP}_{c},
\label{eq:map}
\end{equation}
where \(C\) is the number of classes and \(\mathrm{AP}_{c}\) is the area under the precision–recall curve for class \(c\). In this study, mAP is reported at an IoU threshold of 0.50.

\begin{figure}[htbp]
    \centering
    \includegraphics[width=0.48\textwidth]{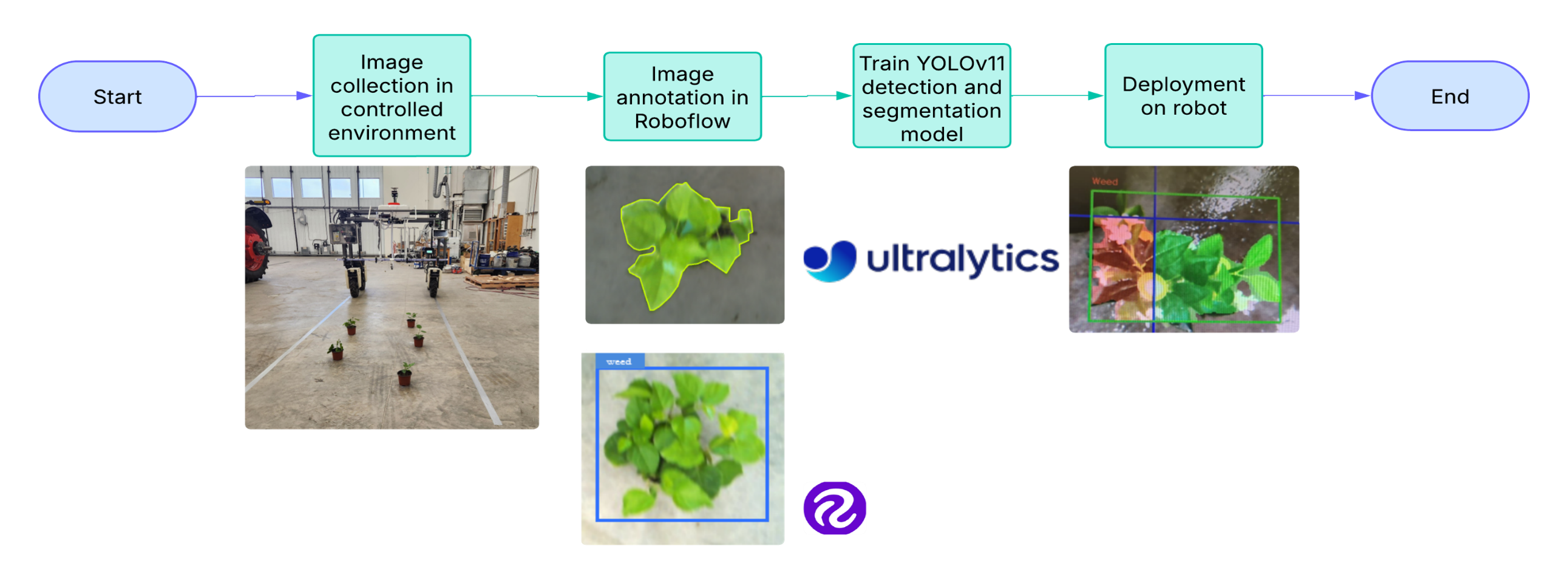}
    \caption{Workflow pipeline from AI model training to deployment on embedded control unit.}
    \label{fig:workflow}
\end{figure}

These metrics were computed on the held-out test set to quantify the effect of the canopy-aware spraying strategy on detection and segmentation accuracy. After the models were trained and evaluated, they were deployed onto an embedded control unit for real time weed detection and spray activation. The complete workflow is illustrated in Figure~\ref{fig:workflow}.

\subsection{Control Logic for AI-driven Canopy-Aware Variable Rate Sprayer System}

\begin{figure*}[t]
    \centering
    \includegraphics[width=\textwidth]{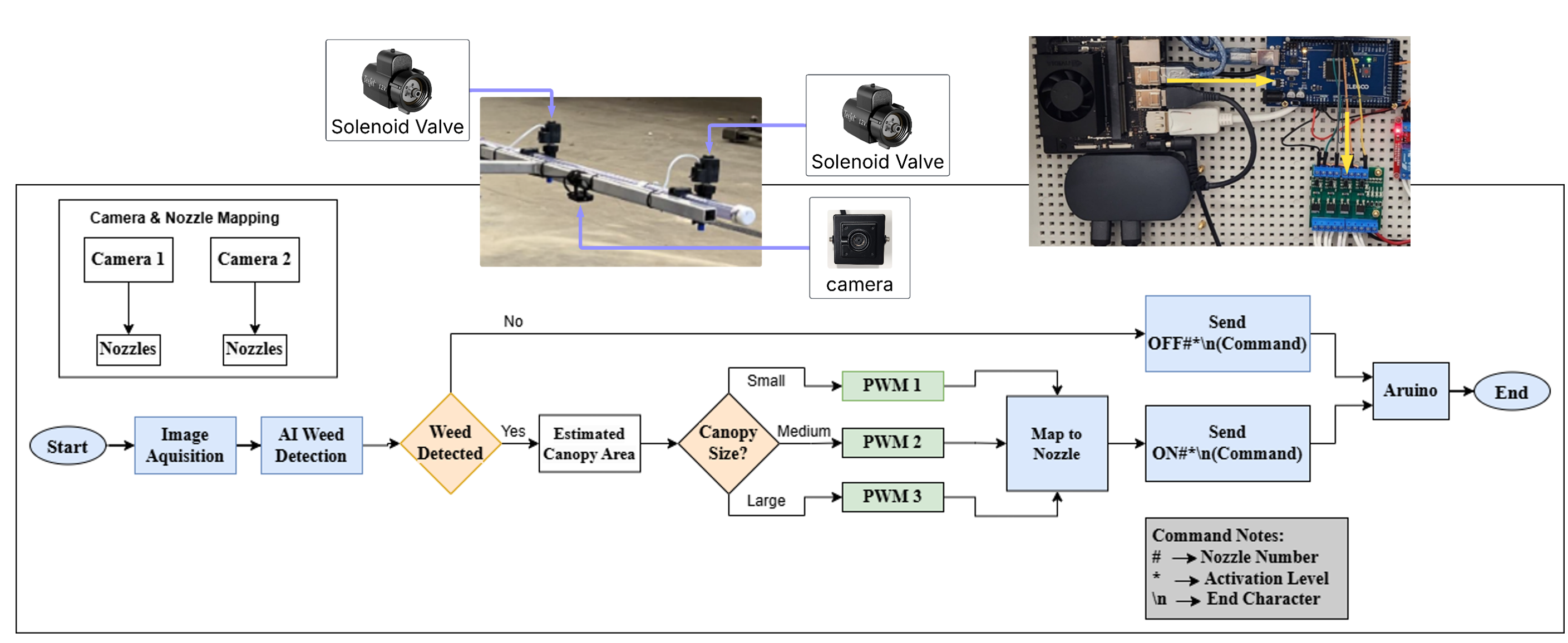}
    \caption{Control flow diagram for AI-based canopy-aware sprayer system. Detected weeds are classified by canopy size and mapped to PWM values to activate specific nozzles.}
    \label{fig:control_logic}
\end{figure*}

The control architecture for the intelligent sprayer system was designed to translate real-time weed detection into targeted spray actuation based on estimated canopy size. As illustrated in Fig.~\ref{fig:control_logic}, the process begins with image acquisition from two front-mounted RGB cameras, each responsible for monitoring a defined set of nozzles. These live images are processed on the Jetson Orin Nano computing unit, which runs a trained YOLO11 and YOLO11n-seg model to detect weed presence and compute canopy area.

If no weed is detected, a command string is sent to the microcontroller to keep all solenoid valves closed: \texttt{OFF\#\textbackslash n}, where \# indicates nozzle number. When a weed is detected, its canopy area is extracted from the segmentation mask and categorized into one of three predefined levels: small, medium, or large. Each category is mapped to a unique PWM duty cycle (PWM1, PWM2, PWM3), corresponding to increasing levels of spray volume. These PWM values are passed along with nozzle ID into a formatted command string: \texttt{ON\#*\textbackslash n}, which is sent to the Arduino Mega via serial communication. The Arduino, in turn, triggers the correct MOSFET channel to actuate the respective solenoid valve. This ensures that each weed receives an appropriately scaled herbicide dose, reducing waste and minimizing overapplication.  

All logic and component mapping were embedded within a real time processing pipeline, ensuring system latency remained under 250 ms from detection to actuation. This rapid response is critical for deployment on mobile robotic platforms in dynamic environments.

\subsection{Spray Pattern Distribution Using Water-Sensitive Paper}

\begin{figure}[htbp]
    \centering
    \includegraphics[width=0.45\textwidth]{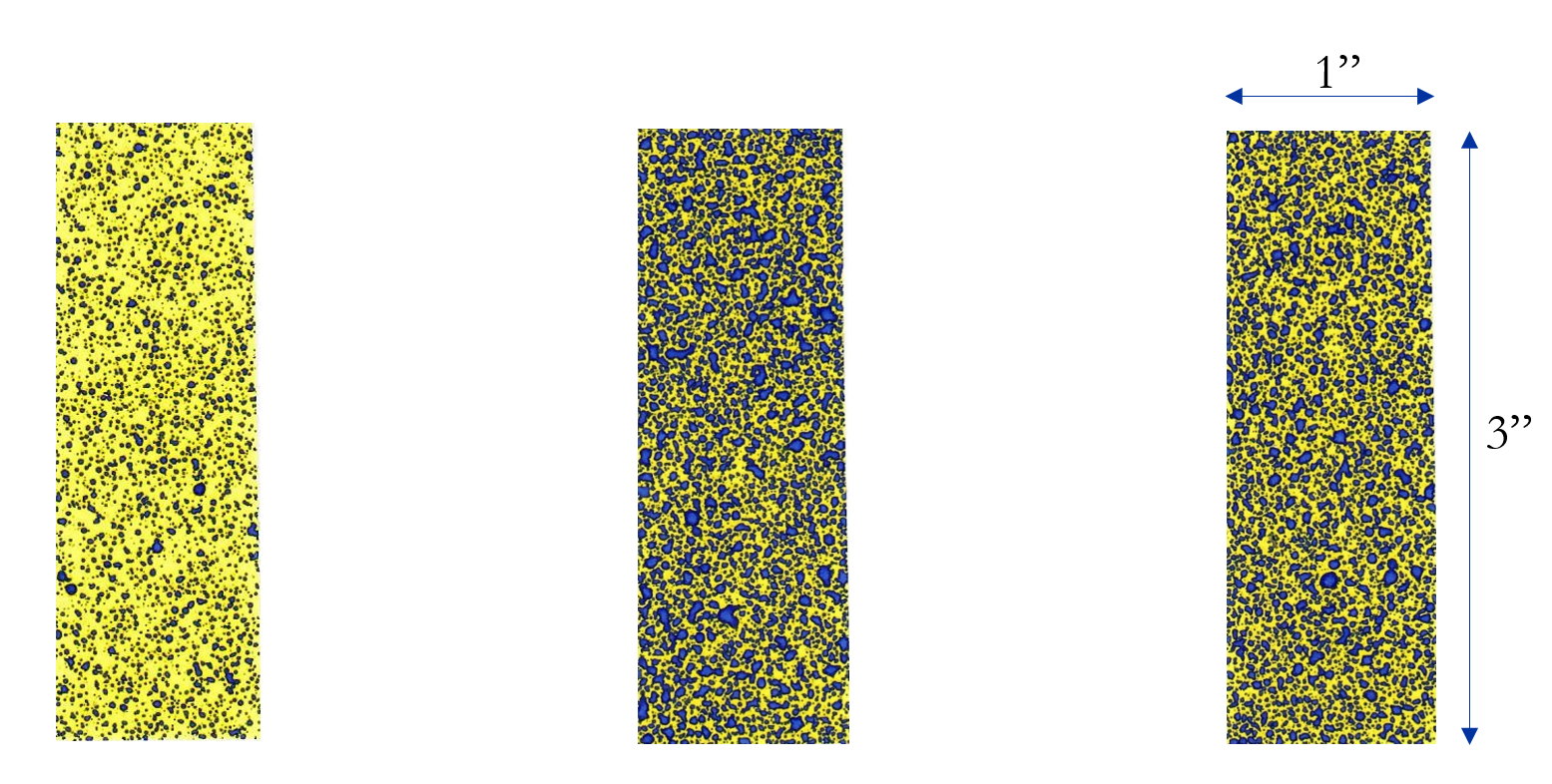}
    \caption{Examples of 1''×3'' water-sensitive paper (WSP) strips showing varying droplet densities used to evaluate spray pattern distribution.}
    \label{fig:wsp_example}
\end{figure}

To evaluate the spray distribution uniformity and coverage efficiency of the AI-driven variable-rate sprayer system, water-sensitive papers (WSPs) were used as the evaluation medium. These yellow chromatic papers (1"×3", Fig.~\ref{fig:wsp_example}) change color upon contact with water, turning blue to indicate droplet impact. This visual transformation allows for precise identification and quantification of droplet characteristics such as size, density, and spatial distribution \cite{foque2010optimization,martin2019effect}.

Each WSP strip was mounted vertically at the height of the plant canopy and exposed to spray treatments for three canopy scenarios: small, medium, and large. The resulting stained strips were photographed and analyzed using a custom Python script that handled image segmentation and droplet detection. The script used color thresholding in the HSV color space, adaptive Gaussian thresholding, and morphological filtering to isolate individual droplets from the background.

Spray coverage percentage was calculated as the ratio of stained area (\(A_{\text{stained}}\)) to total WSP area (\(A_{\text{total}}\)):
\begin{equation}
\text{Spray Coverage (\%)} = \left( \frac{A_{\text{stained}}}{A_{\text{total}}} \right) \times 100
\label{eq:spray_coverage}
\end{equation}

To assess how evenly droplets were distributed, we used kernel density estimation (KDE) to create heatmaps that show high and low concentrations of spray. These visualizations helped compare nozzle activation patterns across canopy sizes and test runs.

For droplet sizing, we calculated the average (\(\mu\)), median, and spread (\(\sigma\)) of the droplet areas:
\begin{equation}
\mu = \frac{1}{N} \sum_{i=1}^{N} d_i
\label{eq:mean_diameter}
\end{equation}

\begin{equation}
\sigma = \sqrt{ \frac{1}{N} \sum_{i=1}^{N} (d_i - \mu)^2 }
\label{eq:std_dev}
\end{equation}

Here, \(d_i\) is the area of the \(i^{th}\) droplet, and \(N\) is the total number of detected droplets.

\begin{figure}[htbp]
    \centering
    \includegraphics[width=0.45\textwidth]{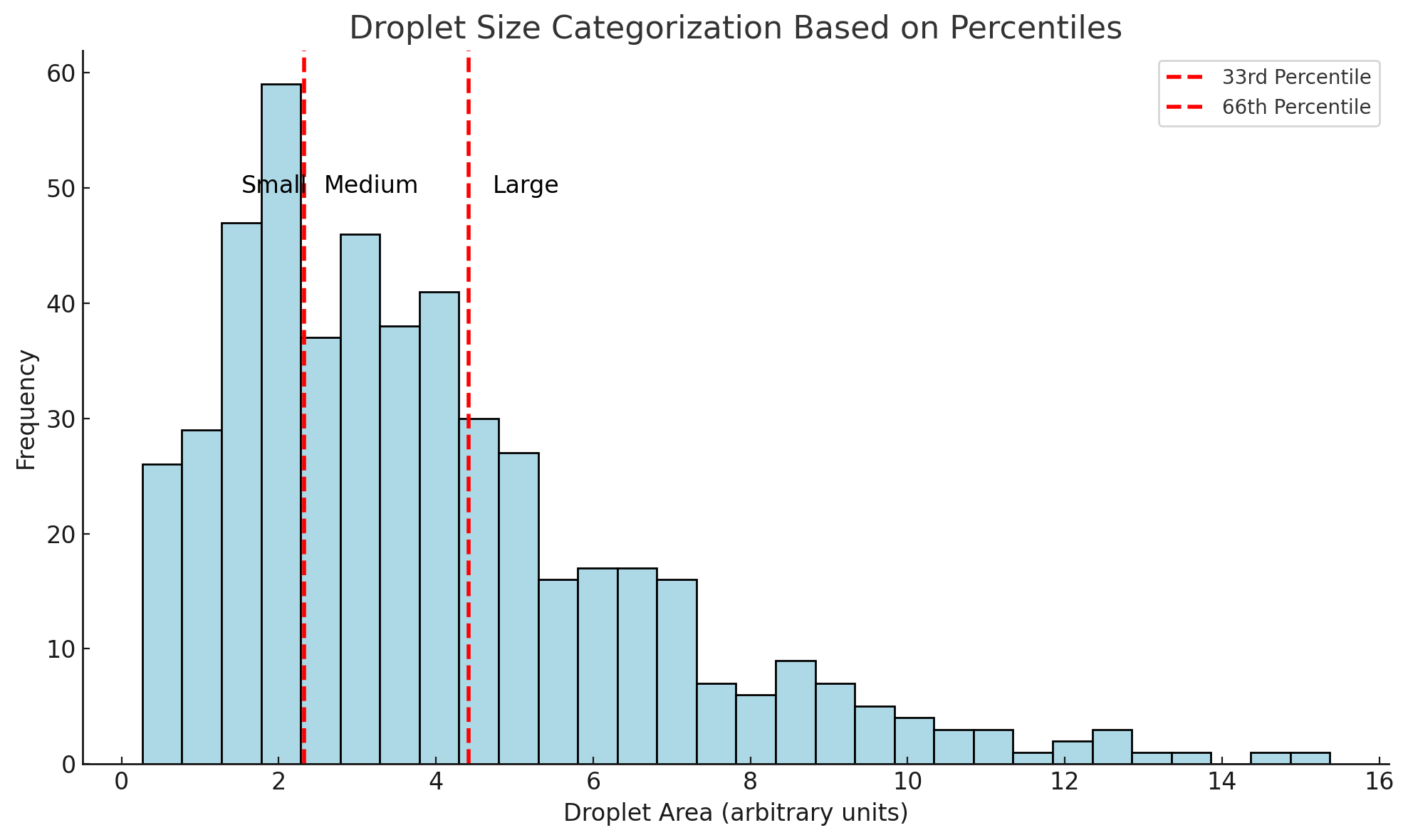}
    \caption{Histogram showing how spray droplets were categorized into small, medium and large based upon percentiles.}
    \label{fig:dropletpercentile}
\end{figure}

Droplets were sorted into three categories:
\begin{itemize}
    \item Small: \( \text{area} < P_{33} \)
    \item Medium: \( P_{33} \leq \text{area} < P_{66} \)
    \item Large: \( \text{area} \geq P_{66} \)
\end{itemize}
where \(P_{33}\) and \(P_{66}\) are the 33rd and 66th percentiles of all droplet areas Fig.~\ref{fig:dropletpercentile}. Counts for each group were used to compare droplet distribution across trials and nozzle settings Fig.~\ref{fig:dropletpercentile}.

We also calculated the uniformity of spraying using the coefficient of variation (CV), and assessed drift by comparing the stained area in the central region of the WSP to the sides.

\section{Results and Discussion}
\subsection{YOLO11n Weed Detection Model}
\label{subsec:yolo11_results}

\begin{figure}[htbp]
    \centering
    \begin{subfigure}[b]{0.48\textwidth}
        \centering
        \includegraphics[width=\linewidth ]{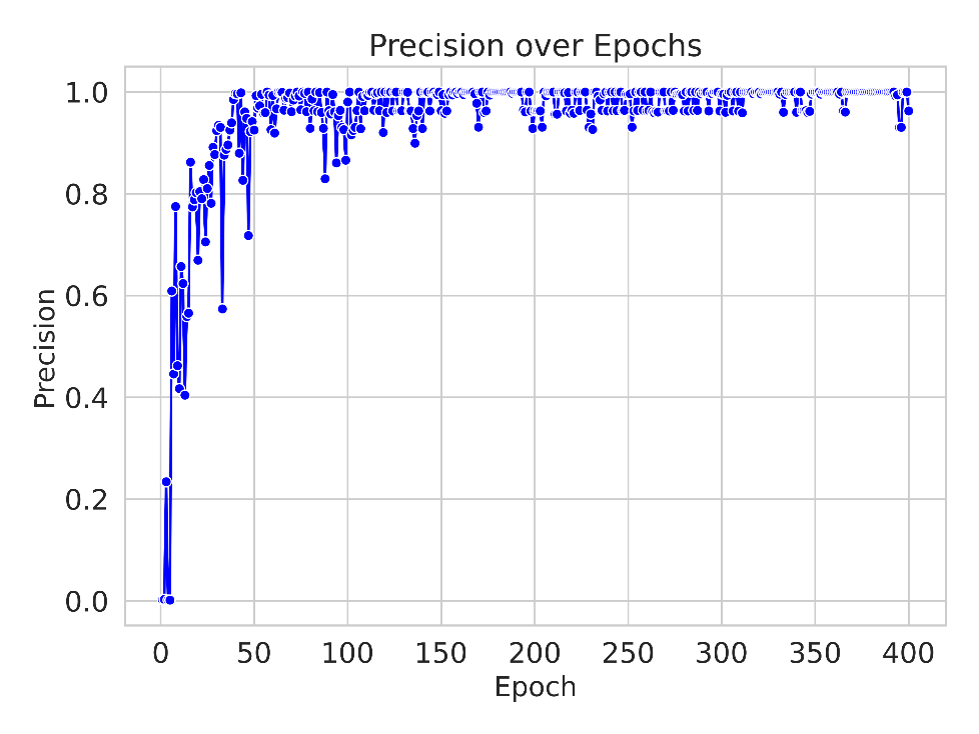}
        \caption{Precision}
    \end{subfigure}
    \hfill
    \begin{subfigure}[b]{0.48\textwidth}
        \centering
        \includegraphics[width=\linewidth]{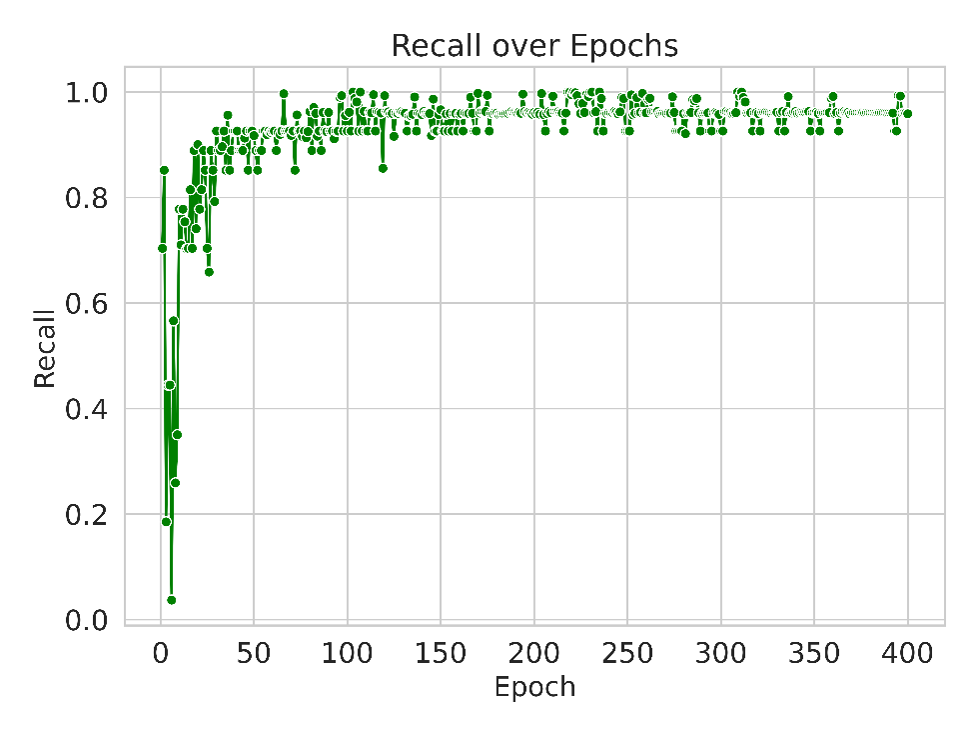}
        \caption{Recall}
    \end{subfigure}
    \hfill
    \begin{subfigure}[b]{0.48\textwidth}
        \centering
        \includegraphics[width=\linewidth]{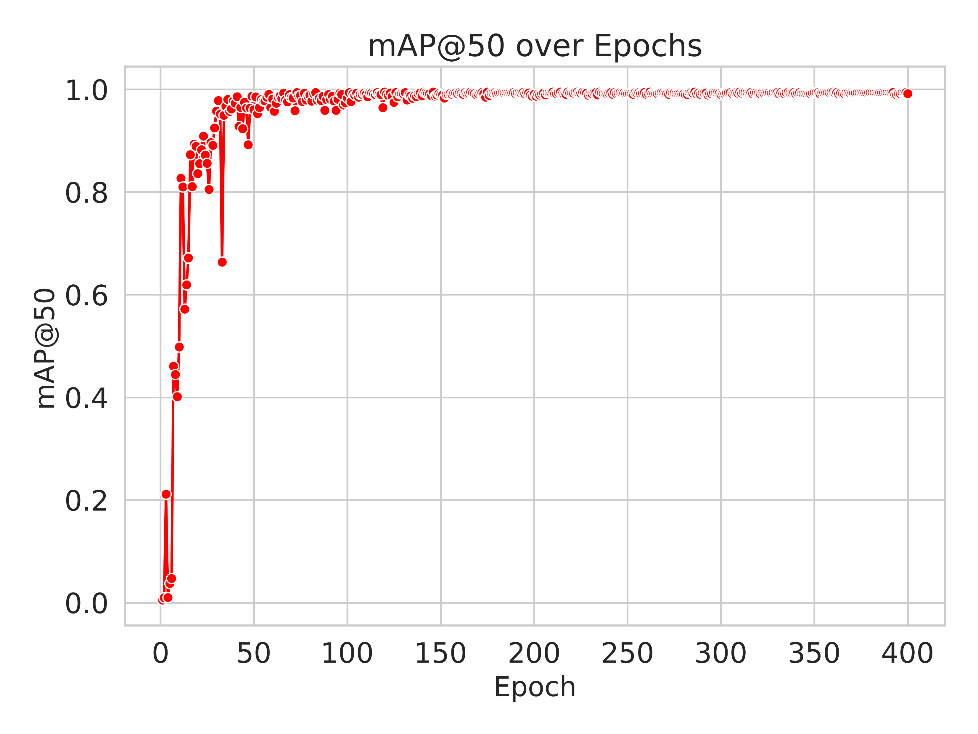}
        \caption{mAP@0.50}
    \end{subfigure}
    \caption{Training curves for YOLO11n over 400~epochs.}
    \label{fig:train_curves}
\end{figure}

Figure~\ref{fig:train_curves} shows the training curves for precision, recall, and mAP@0.50 over 400 epochs.  Model performance improved rapidly during the first 50 epochs, after which all three metrics plateaued.  Precision exceeded 0.95 from epoch~70 onward and stabilised near~0.99, while recall converged slightly below~1.0.  The mAP@0.50 metric displayed a similar trend, reaching~0.98 after roughly 80~epochs and remaining constant thereafter, indicating that additional training yielded only marginal gains in detection accuracy.

\begin{figure}[htbp]
    \centering
    \includegraphics[width=0.48\textwidth]{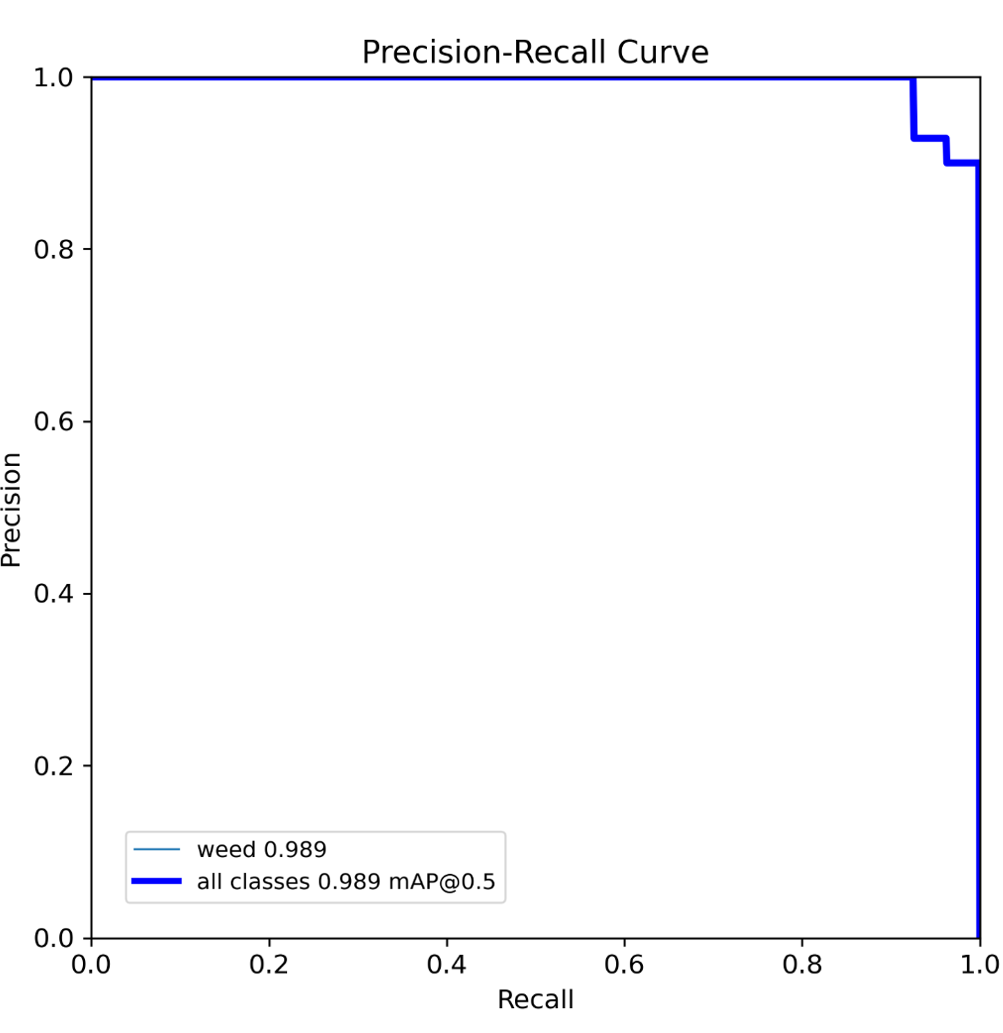}
    \caption{Precision–recall curve for the weed class (test set)}
    \label{fig:pr_curve}
\end{figure}

\begin{figure}[htbp]
    \centering
    \includegraphics[width=0.48\textwidth]{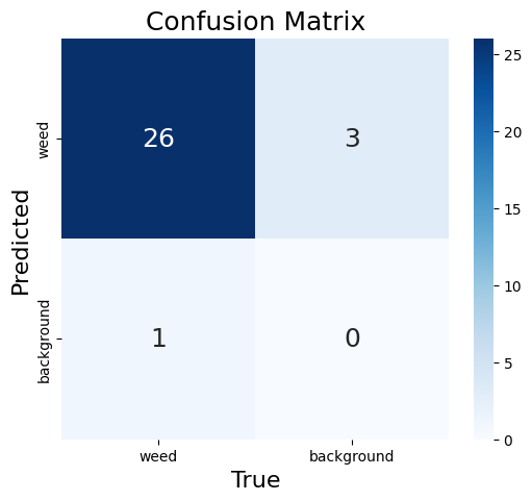}
    \caption{Confusion matrix for YOLO11n on the test set.}
    \label{fig:conf_mat}
\end{figure}

The quality of the final model on the held-out test set is summarised by the precision–recall (PR) curve in Fig.~\ref{fig:pr_curve} and the confusion matrix in Fig.~\ref{fig:conf_mat}.  The PR curve encloses an area of~0.989, confirming a near-ideal trade-off between precision and recall.  The confusion matrix reveals 26 true positives, three false positives, and a single false negative for the weed class, resulting in a class-level precision of 0.897 and recall of 0.963 (Eqs.~\ref{eq:precision}–\ref{eq:recall}).  Although precision is slightly lower than the plateau observed during training, the overall detection reliability remains high and is sufficient for downstream spray-actuation decisions.

Compared with recent studies on vision-based weed detection that report mAP values between 0.80 and 0.95 for similar datasets, the YOLO11n model demonstrates competitive or superior performance despite its compact architecture and modest training set of 443 images.  The rapid convergence of all metrics underscores the benefit of data augmentation and the suitability of the YOLO11 backbone for real-time field deployment on embedded hardware.

\subsection{YOLO11n-seg Weed Segmentation Model}

To accurately delineate weed canopy areas for precision variable-rate spraying, a lightweight YOLO11n-seg segmentation model was trained. The training dataset consisted of 1443 original images of 15 potted weed plants with varying canopy sizes, collected during the indoor trials. These were augmented using a range of techniques such as rotation, flipping, brightness variation, and scaling, which allowed expansion of the training, validation, and test sets while ensuring model generalizability.

\begin{figure}[htbp]
    \centering
    \begin{subfigure}[b]{0.42\textwidth}
        \centering
        \includegraphics[width=\linewidth, height=5cm]{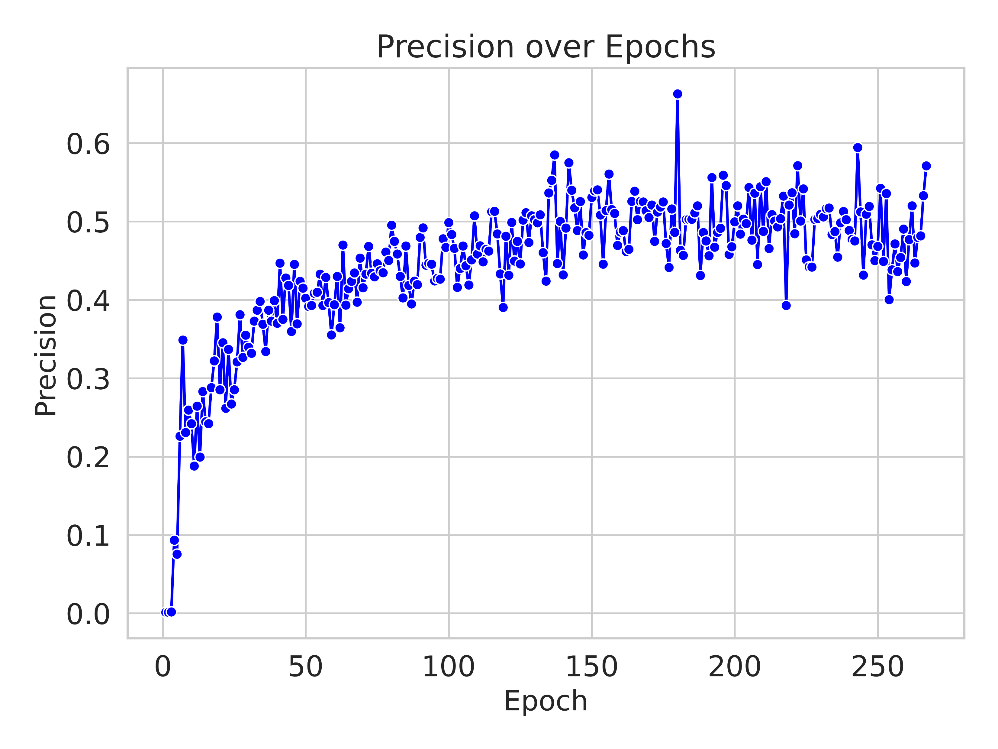}
        \caption{Precision over epochs for YOLO11n-seg model}
    \end{subfigure}
    \hfill
    \begin{subfigure}[b]{0.42\textwidth}
        \centering
        \includegraphics[width=\linewidth,height=5cm]{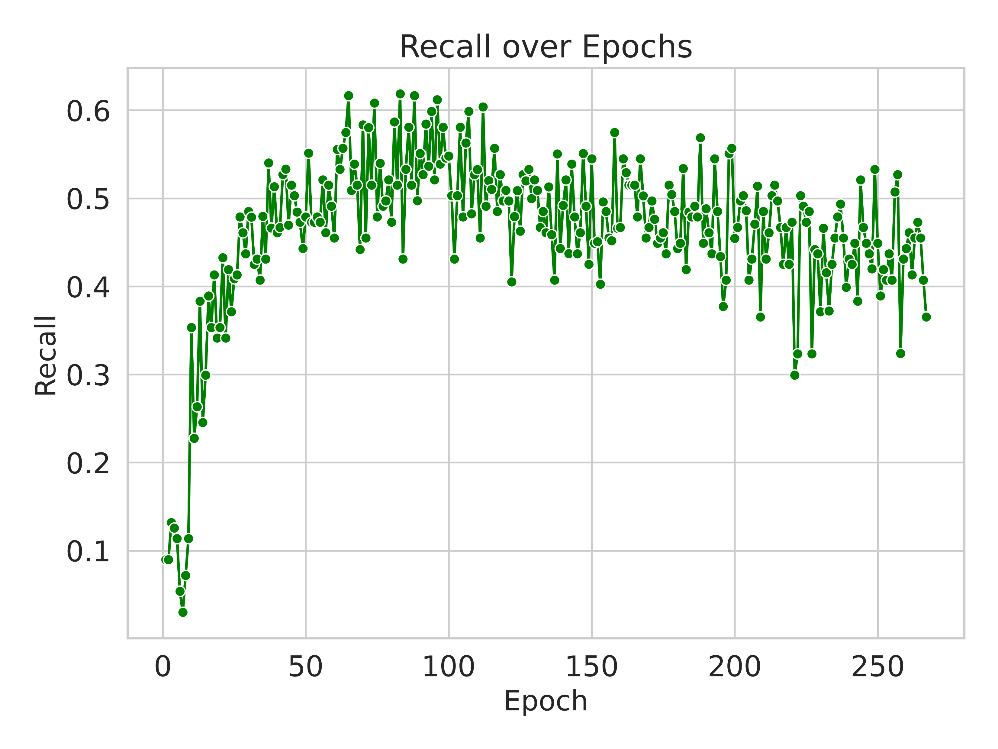}
        \caption{Recall over epochs for YOLO11n-seg model}
    \end{subfigure}
    \hfill
    \begin{subfigure}[b]{0.42\textwidth}
        \centering
        \includegraphics[width=\linewidth,height=5cm]{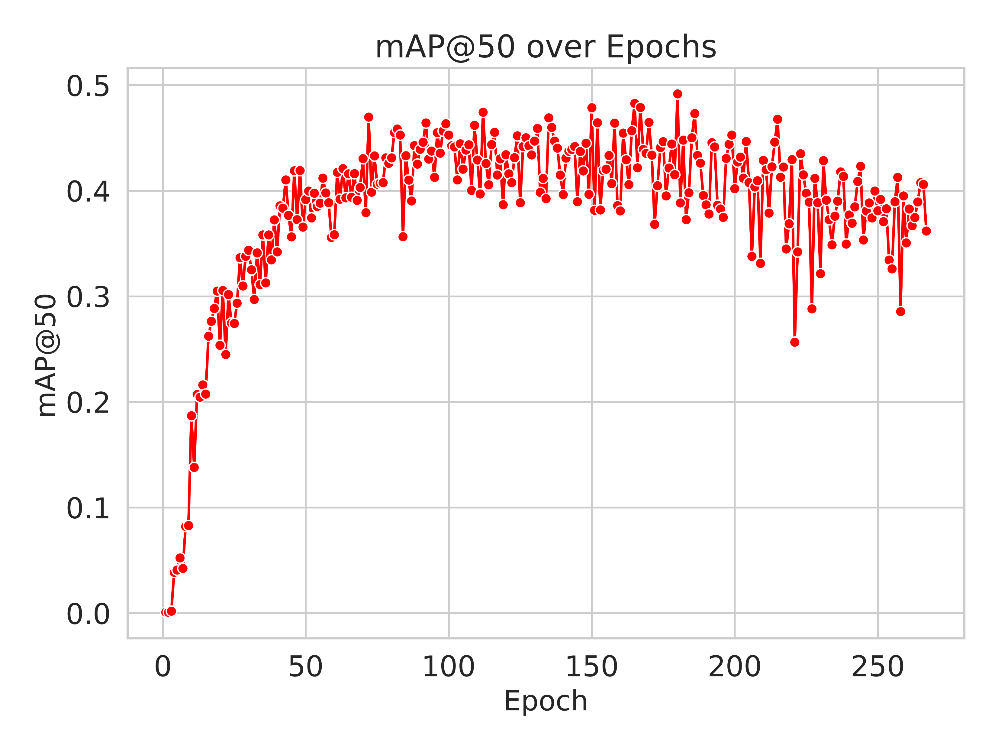}
        \caption{mAP@50 over epochs for YOLO11n-seg model}
    \end{subfigure}
    \caption{Training curves for YOLO11n over 400~epochs.}
    \label{fig:train_curves_yolon-seg}
\end{figure}

The YOLO11n-seg model was trained for 270 epochs with a batch size of 16 and an initial learning rate of 0.001. Model performance was evaluated based on key metrics such as precision, recall, and mean Average Precision at 0.5 IoU (mAP@50). Figures~\ref{fig:train_curves_yolon-seg}, \ref{fig:train_curves_yolon-seg}, and \ref{fig:train_curves_yolon-seg} illustrate the evolution of these metrics over the training epochs. Precision and recall stabilized after around 150 epochs, with the final mAP@50 value reaching approximately 0.48.

\begin{figure}[htbp]
    \centering
    \includegraphics[width=0.45\textwidth]{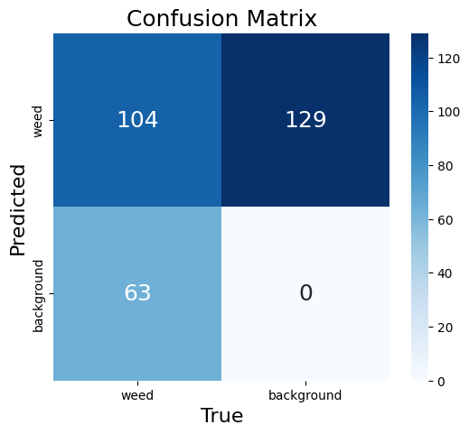}
    \caption{Confusion matrix showing predicted vs. true weed and background pixels.}
    \label{fig:yolo11nseg_confmat}
\end{figure}

A detailed precision-recall curve (Figure~\ref{fig:train_curves_yolon-seg}) and confusion matrix (Figure~\ref{fig:yolo11nseg_confmat}) were generated to evaluate the classification performance. The model achieved a precision of 0.48 for the "weed" class, with several false positives and false negatives observed, primarily due to partial occlusions and variability in canopy morphology.

The segmentation masks generated by the YOLO11n-seg model were used as inputs for real-time spray control logic. These masks enabled pixel-level estimation of canopy area, which guided the nozzle-specific PWM control commands for activating the solenoid valves.

\begin{figure}[htbp]
    \centering
    \includegraphics[width=0.45\textwidth]{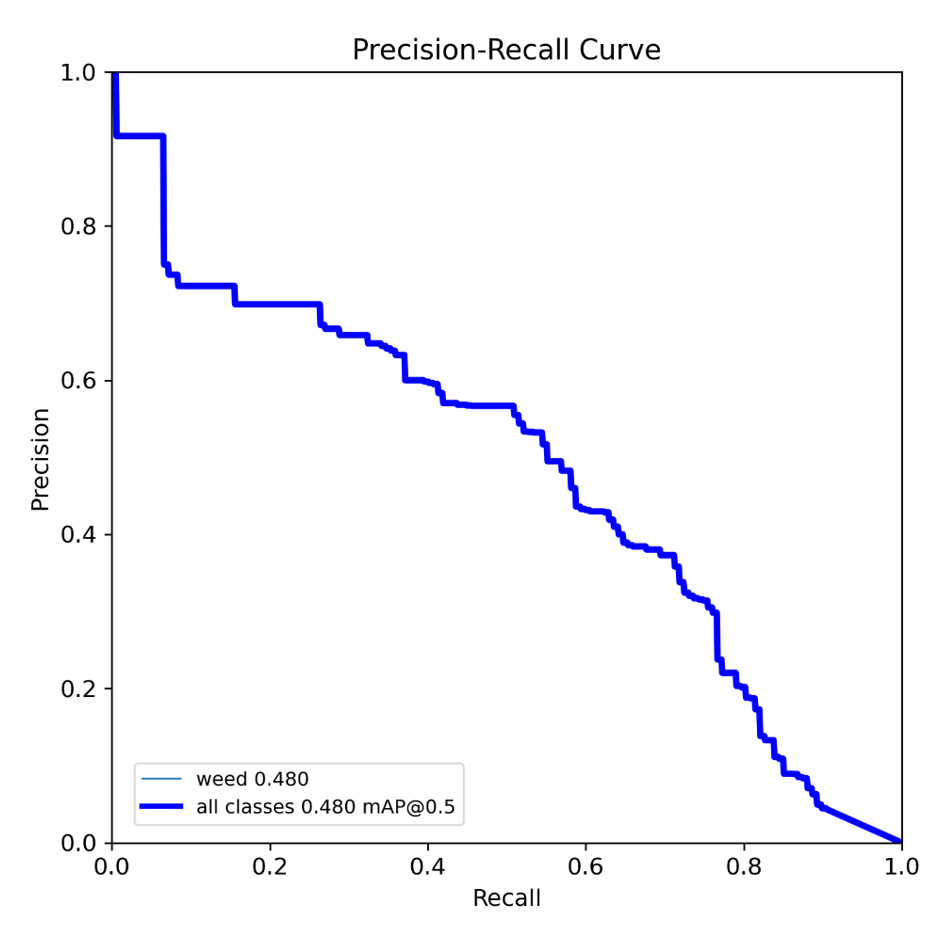}
    \caption{Precision-Recall curve showing detection performance for YOLOv11-seg model.}
    \label{fig:yolo11nseg_prcurve}
\end{figure}

\subsection{Droplet Distribution and Spray Percentage Coverage on Water Sensitive Paper}

To validate the canopy aware functionality of the AI-driven variable rate sprayer system, water sensitive papers (WSPs) were used to assess how well the system adjusted spray output based on canopy size. These 1''×3'' yellow chromatic papers were placed at plant height across three canopy categories: small, medium, and large. Upon contact with spray droplets, the WSPs turned blue, providing a visual and quantifiable record of droplet deposition patterns.

\begin{figure}[htbp]
    \centering
    \includegraphics[width=0.48\textwidth]{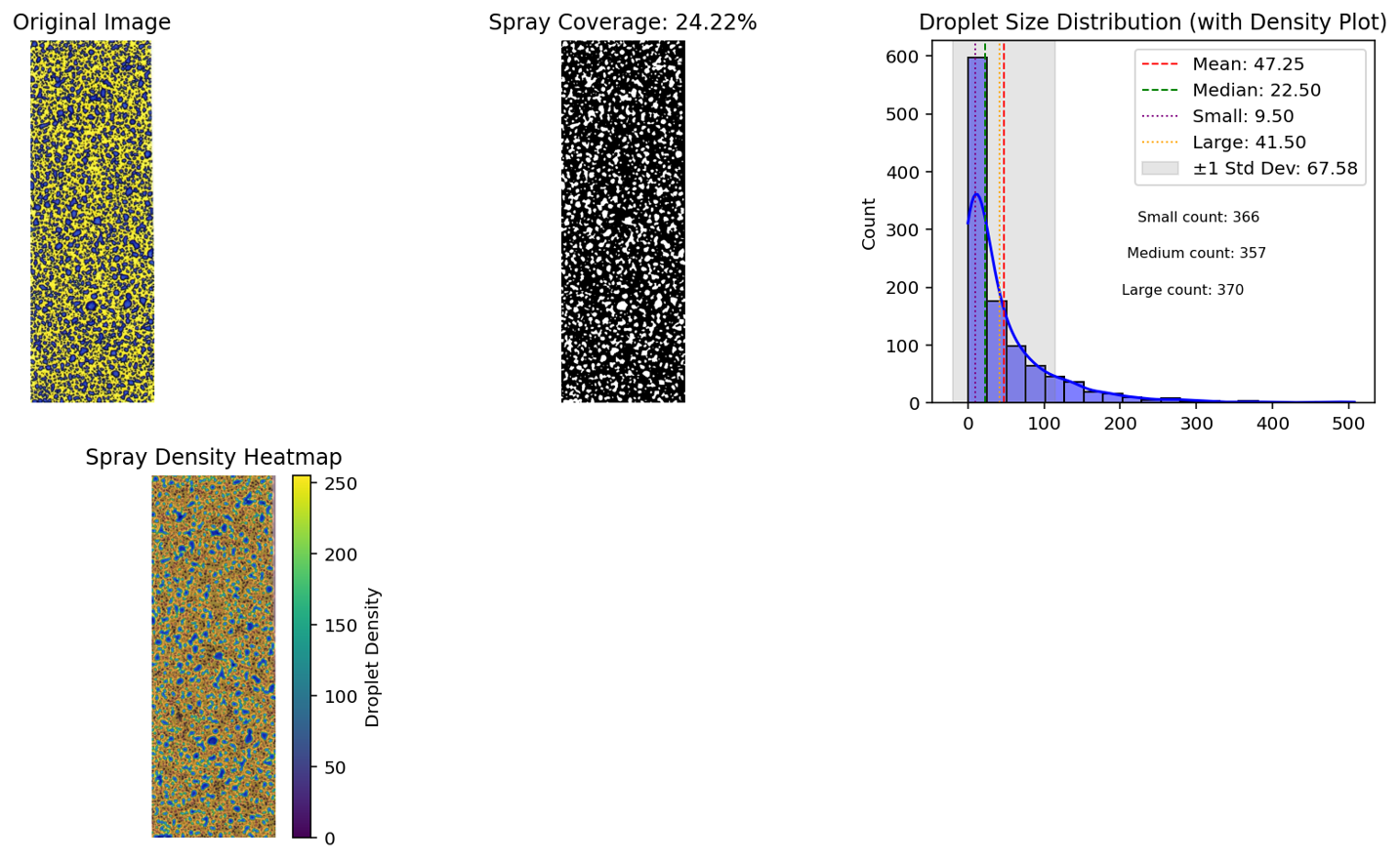}
    \caption{Spray analysis of a water-sensitive paper: (Top row) original image, binary spray mask (24.22\% coverage), droplet size histogram with kernel density plot; (Bottom) droplet density heatmap.}
    \label{fig:coverage_density}
\end{figure}

Figure~\ref{fig:coverage_density} illustrates the analysis pipeline for a representative WSP strip, including the original image, binary thresholded spray coverage mask, droplet size distribution, and a droplet density heatmap. The sprayer achieved an average spray coverage of 24.22\%, indicating partial but targeted application. Most droplets were smaller than 100~$\mu$m, with a mean size of 47.25~$\mu$m and a standard deviation of 67.58~$\mu$m. Droplets were grouped into three categories based on area percentiles: small ($<$9.5~$\mu$m), medium (9.5–41.5~$\mu$m), and large ($>$41.5~$\mu$m), with nearly balanced counts of 366, 357, and 370 droplets, respectively. This balance supports the system’s ability to maintain effective droplet size distribution for coverage without over saturation.

\begin{figure}[htbp]
    \centering
    \includegraphics[width=0.48\textwidth]{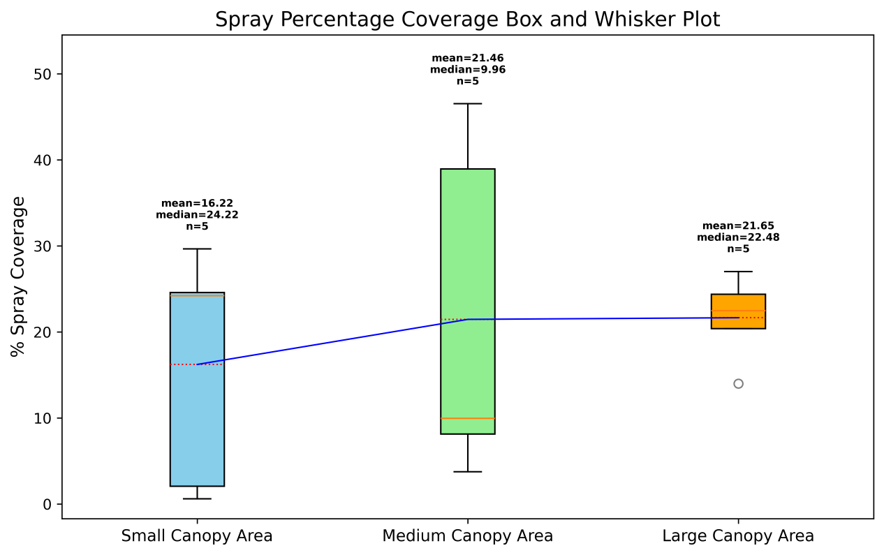}
    \caption{Box-and-whisker plot comparing spray coverage across canopy sizes (small, medium, large), with $n=5$ replicates per category.}
    \label{fig:boxplot_spray}
\end{figure}

To examine how well the system modulated spray rate in response to detected canopy size, box-and-whisker plots were generated for each canopy category using five WSP replicates (Fig.~\ref{fig:boxplot_spray}). The average spray coverage progressively increased with canopy size 16.22\% for small, 21.46\% for medium, and 21.65\% for large canopy zones. A similar trend was observed in the median coverage values: 9.96\%, 24.22\%, and 22.48\% for small, medium, and large canopies, respectively. This indicates that the system effectively responded to real time canopy detection, adjusting nozzle activation and spray output accordingly. Variability observed in the medium canopy group may be due to more irregular plant shapes or occasional misclassification in segmentation.

These findings demonstrate that the vision guided sprayer could dynamically and consistently adjust spray output based on real time canopy information. While larger canopies received higher and more uniform spray coverage, smaller canopies were appropriately sprayed with reduced volume, minimizing overspray and improving efficiency. These results highlight the potential of AI integrated sprayer systems for precise, input optimized field applications.

\FloatBarrier  
\section{Conclusion and Future Work}

This study presents the design, development, and initial testing of a vision guided, AI-driven variable rate sprayer system capable of real time weed detection and site specific herbicide application. The system combines a custom trained YOLO11n-seg semantic segmentation model with a Jetson Orin Nano combined witrh Arduino Uno to detect canopy presence and dynamically control nozzle activation. Spray pattern analysis using water sensitive papers confirmed the system’s ability to adjust spray rates based on canopy size, achieving targeted delivery with an average spray coverage of 24.22\% and a balanced droplet size distribution across small, medium, and large plant zones.

The modular sprayer platform built using aluminum extrusions, low-cost Arducam IMX219 cameras, and TeeJet components including QJ22187 nozzle bodies and 115880-2-12 solenoid valves was designed with scalability and flexibility in mind. The system performed well under controlled indoor conditions, demonstrating its potential for resource efficient weed management. Model evaluation metrics showed promising results with mAP@50 of 0.48, precision of 0.55, and recall of 0.52. While these figures suggest effective detection performance, further improvements are needed to reduce false detections and improve segmentation robustness under varied lighting and occlusion scenarios.

Our future work will involve both indoor and field trials using three prevalent weed species in South Dakota: waterhemp (\textit{Amaranthus tuberculatus} L.), kochia (\textit{Bassia scoparia} L.), and foxtail (\textit{Setaria spp.}). These trials will help test the adaptability of the system across realistic farm environments and variable plant architectures. Enhancements will also focus on refining segmentation accuracy, reducing nozzle actuation latency, and streamlining image acquisition for faster processing. Long term objectives include integrating prescription mapping tools and automating full field spraying workflows. Overall, this work lays a solid foundation for the development of smart, sustainable spraying technologies that reduce chemical usage, enhance targeting precision, and contribute to responsible agricultural practices.

\section*{Acknowledgment}

This research was supported by the United State Department of Agriculture-National Institute of Food \& Agriculture (USDA-NIFA) Multistate Hatch. The authors acknowledge the help of MVOS lab group member Kasish Siwakoti and Brian Langum for assisting with experiments and indoor trials.

\bibliographystyle{IEEEtran}

\end{document}